\documentclass[12pt]{article}
\usepackage[centertags]{amsmath}
\usepackage{amsfonts}
\usepackage{amssymb}
\usepackage{latexsym}
\usepackage{amsthm}
\usepackage{newlfont}
\usepackage{graphicx}
\usepackage{listings}
\usepackage{booktabs}
\usepackage{abstract}
\newlength{\defbaselineskip}
\newcommand{\setlinespacing}[1]%
           {\setlength{\baselineskip}{#1 \defbaselineskip}}

\newcommand{\actaqed}{\hfill $\actabox$}
{\medskip\noindent \textit{Proof of #1. }}%
{\actaqed \medskip}

\def\D{{\mathcal D}}
\def\A{{\mathcal A}}
\def\H{{\mathcal H}}

\def \<{\langle}
\def\>{\rangle}

\def \e{\epsilon}
\def \de{\delta}

\def \sp{\operatorname{span}}

\def \sign{\operatorname{sign}}
\def \exp{\operatorname{exp}}

\def\la{\lambda}

\newtheorem{Theorem}{Theorem}[section]
\newtheorem{Lemma}{Lemma}[section]
\newtheorem{Definition}{Definition}[section]

\newtheorem{Remark}{Remark}[section]

\newtheorem{Corollary}{Corollary}[section]
\numberwithin{equation}{section}

\begin{document}
\title{{Sparse approximation and recovery by greedy algorithms in Banach spaces}\thanks{\it Math Subject Classifications.
primary:  41A65; secondary: 41A25, 41A46, 46B20.}}
\author{V. Temlyakov \thanks{ University of South Carolina and Steklov Institute of Mathematics. Research was supported by NSF grant DMS-1160841 }} \maketitle
\begin{abstract}
{We study sparse approximation by greedy algorithms. We prove the Lebesgue-type inequalities for the Weak Chebyshev Greedy Algorithm (WCGA),  a generalization of the Weak Orthogonal Matching Pursuit to the case of a Banach space. The main novelty of these results is a Banach space setting instead of a Hilbert space setting. The results are proved for redundant dictionaries satisfying certain conditions. Then we apply these general results to the case of bases. In particular, we prove that the WCGA provides almost optimal sparse approximation for the trigonometric system in $L_p$, $2\le p<\infty$. }
\end{abstract}

\section{Introduction}

 This paper is devoted to theoretical aspects of sparse approximation.  The main motivation for the study of sparse approximation is that many real world signals can be well approximated by sparse ones.   Sparse approximation
automatically implies a need for nonlinear approximation, in particular, for greedy approximation. We give a brief description of a sparse approximation problem. In a general setting we are working in a Banach space $X$ with a redundant system of elements $\D$ (dictionary $\D$). There is a solid justification of importance of a Banach space setting in numerical analysis in general and in sparse approximation in particular (see, for instance, \cite{Tbook}, Preface, and \cite{ST}). 
An element (function, signal) $f\in X$ is said to be $K$-sparse with respect to $\D$ if
it has a representation $f=\sum_{i=1}^Kx_ig_i$,   $g_i\in \D$, $i=1,\dots,K$. The set of all $K$-sparse elements is denoted by $\Sigma_K(\D)$. For a given element $f_0$ we introduce the error of best $m$-term approximation
$$
\sigma_m(f_0,\D) := \inf_{f\in\Sigma_m(\D)} \|f_0-f\|.
$$
We are interested in the following fundamental problem of sparse approximation. 

{\bf Problem.} How to design a practical algorithm that builds sparse approximations comparable to best $m$-term approximations? 
 
We demonstrate in this paper that the Weak Chebyshev Greedy Algorithm (WCGA) which we define momentarily is a solution to the above problem. 
This paper is devoted to the Banach space setting.
Let $X$ be a real Banach space with norm $\|\cdot\|:=\|\cdot\|_X$. We say that a set of elements (functions) $\D$ from $X$ is a dictionary  if each $g\in \D$ has norm   one ($\|g\|=1$),
and the closure of $\sp \D$ is $X$.
For a nonzero element $g\in X$ we let $F_g$ denote a norming (peak) functional for $g$:
$$
\|F_g\|_{X^*} =1,\qquad F_g(g) =\|g\|_X.
$$
The existence of such a functional is guaranteed by the Hahn-Banach theorem.

Let
$\tau := \{t_k\}_{k=1}^\infty$ be a given weakness sequence of  nonnegative numbers $t_k \le 1$, $k=1,\dots$. We define the Weak Chebyshev Greedy Algorithm (WCGA) (see \cite{T15}) as a generalization for Banach spaces of the Weak Orthogonal Matching Pursuit (WOMP). In a Hilbert space the WCGA coincides with the WOMP. The WOPM is very popular in signal processing, in particular, in compressed sensing.  We study in detail the WCGA in this paper.

 {\bf Weak Chebyshev Greedy Algorithm (WCGA).}
Let $f_0$ be given. Then for each $m\ge 1$ we have the following inductive definition.

(1) $\varphi_m :=\varphi^{c,\tau}_m \in \D$ is any element satisfying
$$
|F_{f_{m-1}}(\varphi_m)| \ge t_m\sup_{g\in\D}  | F_{f_{m-1}}(g)|.
$$

(2) Define
$$
\Phi_m := \Phi^\tau_m := \sp \{\varphi_j\}_{j=1}^m,
$$
and define $G_m := G_m^{c,\tau}$ to be the best approximant to $f_0$ from $\Phi_m$.

(3) Let
$$
f_m := f^{c,\tau}_m := f_0-G_m.
$$
In this paper we only consider the case when $t_k=t\in(0,1]$, $k=1,2,\dots$.

The trigonometric system is a classical system that is known to be difficult to study. In this paper we study among other problems the problem of nonlinear sparse approximation with respect to it. Let  ${\mathcal R}{\mathcal T}$ denote the real trigonometric system 
$1,\sin 2\pi x,\cos 2\pi x, \dots$ on $[0,1]$ and let ${\mathcal R}{\mathcal T}_p$ to be its version normalized in $L_p([0,1])$. Denote ${\mathcal R}{\mathcal T}_p^d := {\mathcal R}{\mathcal T}_p\times\cdots\times {\mathcal R}{\mathcal T}_p$ the $d$-variate trigonometric system. We need to consider the real trigonometric system because the algorithm WCGA is well studied for the real Banach space. In order to illustrate performance of the WCGA we   discuss in this section  the above mentioned problem for the trigonometric system. 
There is a natural algorithm, the Thresholding Greedy Algorithm (TGA), that can be considered 
for the above problem. We give a definition of the TGA for a general basis $\Psi$.
Let a Banach space $X$, with a normalized basis $\Psi =\{\psi_k\}_{k=1}^\infty$,    be given. We consider the following  greedy algorithm. For a given element $f\in X$ we consider the expansion
\begin{equation}\label{2c1.1}
f = \sum_{k=1}^\infty c_k(f)\psi_k. 
\end{equation}
For an element $f\in X$ we say that a permutation $\rho$ of the positive integers   is  decreasing  if 
\begin{equation}\label{2c1.2}
|c_{k_1}(f) |\ge |c_{k_2}(f) | \ge \dots ,  
\end{equation}
 where $\rho(j)=k_j$, $j=1,2,\dots$, and write $\rho \in D(f)$.
If the   inequalities are strict in (\ref{2c1.2}), then $D(f)$ consists of only one permutation. We define the $m$th greedy approximant of $f$, with regard to the basis $\Psi$ corresponding to a permutation $\rho \in D(f)$, by the formula
$$
 G_m(f,\Psi)  :=G_m(f,\Psi,\rho) := \sum_{j=1}^m c_{k_j}(f)\psi_{k_j}.
$$
The following Lebesgue-type inequality was proved in  
\cite{T10}.
\begin{Theorem}\label{T1.1}  
For each $f \in L_p([0,1]^d)$ we have
$$
\|f - G_m(f,{\mathcal R\mathcal T}^d_p)\|_p \le C(d)m^{h(p)}\sigma_m(f,{\mathcal R\mathcal T}^d_p)_p , \quad 1 \le p \le \infty,
$$
where $h(p) := |1/2-1/p|.$
\end{Theorem}
It was also proved in \cite{T10} that the above inequality is sharp.
\begin{Remark}\label{R1.2} There is a positive absolute constant $C$ such that
 for each $m$ and $1 \le p \le \infty$ there exists a function $f\neq 0$ with the
property
\begin{equation}\label{t4.1}
\|G_m(f,{\mathcal R}{\mathcal T}_p)\|_p \ge Cm^{h(p)}\|f\|_p . 
\end{equation}
\end{Remark}
  Remark \ref{R1.2} shows that the TGA does not work well for the trigonometric system in $L_p$, $p\neq 2$. This leads to a natural attempt to consider some other algorithms that may have some advantages over the TGA in the case of the trigonometric system. In this  paper we discuss the performance of the Weak Chebyshev Greedy Algorithm (WCGA) with respect to the trigonometric system. We prove here the following 
Lebesgue-type inequality for the WCGA (see Example 2 in Section 4).
\begin{Theorem}\label{T1.2} Let $\D$ be the normalized in $L_p$, $2\le p<\infty$, real $d$-variate trigonometric
system. Then    
for any $f_0\in L_p$ the WCGA with weakness parameter $t$ gives
\begin{equation}\label{I1.4}
\|f_{C(t,p,d)m\ln (m+1)}\|_p \le C\sigma_m(f_0,\D)_p .
\end{equation}
\end{Theorem}
The Open Problem 7.1 (p. 91) from \cite{Tsurv} asks if (\ref{I1.4}) holds without an extra 
$\ln (m+1)$ factor. Theorem \ref{T1.2} is the first result on the Lebesgue-type inequalities for the WCGA with respect to the trigonometric system. It provides a progress in solving the above mentioned open problem, but the problem is still open. 

Theorem \ref{T1.2} shows that the WCGA is very well designed for the trigonometric system. We show in Example 1 of Section 4 that an analog of (\ref{I1.4}) holds for 
uniformly bounded orthogonal systems. We note that it is known (see \cite{Tbook}) that the TGA is very well designed for bases $L_p$-equivalent to the Haar basis, $1<p<\infty$.
We discuss performance of the WCGA in more detail in Section 5.

The proof of Theorem \ref{T1.2} uses technique developed  for proving the Lebesgue-type inequalities for redundant dictionaries with special properties. We present these results in 
Sections 2 and 3. These results are an extension of earlier results from \cite{LivTem}. 
In Section 4 we test the power of general results from Section 2 on specific dictionaries, namely, on bases. 
Section 4 provides a number of examples, including the trigonometric system, were the technique from Sections 2 and 3 can be successfully applied. In particular, results from Section 4 demonstrate that the general technique from Sections 2 and 3 provides almost optimal $m$-term approximation results for uniformly bounded orthogonal systems (see Example 1). Example 7 shows that an extra assumption that a uniformly bounded orthogonal  system $\Psi$ is a quasi-greedy basis allows us to improve inequality (\ref{I1.4}):
\begin{equation*} 
\|f_{C(t,p,\Psi)m\ln \ln (m+3)}\|_p \le C\sigma_m(f_0,\Psi)_p .
\end{equation*}

\section{Lebesgue-type inequalities. General results.}
  
   A very important advantage of the WCGA  is its convergence and rate of convergence properties. The WCGA is well defined for all $m$. Moreover, it is known (see \cite{T15} and \cite{Tbook}) that the WCGA with $\tau=\{t\}$ converges for all $f_0$ in all uniformly smooth Banach spaces with respect to any dictionary. That is, when $X$ is a real Banach space and the modulus of smoothness of $X$ is defined as follows
\begin{equation}\label{1.4}
\rho(u):=\frac{1}{2}\sup_{x,y;\|x\|= \|y\|=1}\left|\|x+uy\|+\|x-uy\|-2\right|,
\end{equation}
then the uniformly smooth Banach space is the one with $\rho(u)/u\to 0$ when $u\to 0$.

We discuss here the Lebesgue-type inequalities for the WCGA with $\tau =\{t\}$, $t\in(0,1]$. For notational convenience we consider here a countable dictionary $\D=\{g_i\}_{i=1}^\infty$. The following assumptions {\bf A1} and {\bf A2} were used in \cite{LivTem}.
 For a given $f_0$ let sparse element (signal)
 $$
 f:=f^\e=\sum_{i\in T}x_ig_i
 $$
 be such that $\|f_0-f^\e\|\le \e$ and $|T|=K$. For $A\subset T$ denote
 $$
 f_A:=f_A^\e := \sum_{i\in A}x_ig_i.
 $$
 
  {\bf A1.} We say that $f=\sum_{i\in T}x_ig_i$ satisfies the Nikol'skii-type $\ell_1X$ inequality with parameter $r$ if
 \begin{equation}\label{C1}
 \sum_{i\in A} |x_i| \le C_1|A|^{r}\|f_A\|,\quad A\subset T,\quad r\ge 1/2.
 \end{equation}
 We say that a dictionary $\D$ has the Nikol'skii-type $\ell_1X$ property with parameters $K$, $r$   if any $K$-sparse element satisfies the Nikol'skii-type
 $\ell_1X$ inequality with parameter $r$.

{\bf A2.}  We say that $f=\sum_{i\in T}x_ig_i$ has incoherence property with parameters $D$ and $U$ if for any $A\subset T$ and any $\Lambda$ such that $A\cap \Lambda =\emptyset$, $|A|+|\Lambda| \le D$ we have for any $\{c_i\}$
\begin{equation}\label{C2}
\|f_A-\sum_{i\in\Lambda}c_ig_i\|\ge U^{-1}\|f_A\|.
\end{equation}
We say that a dictionary $\D$ is $(K,D)$-unconditional with a constant $U$ if for any $f=\sum_{i\in T}x_ig_i$ with
$|T|\le K$ inequality (\ref{C2}) holds.

The term {\it unconditional} in {\bf A2} is justified by the following remark. The above definition of $(K,D)$-unconditional dictionary is equivalent to the following definition. Let $\D$ be such that any subsystem of $D$ distinct elements $e_1,\dots,e_D$ from $\D$ is linearly independent and for any $A\subset [1,D]$ with $|A|\le K$ and any coefficients $\{c_i\}$ we have
$$
\|\sum_{i\in A}c_ie_i\| \le U\|\sum_{i=1}^Dc_ie_i\|.
$$

It is convenient for us to use the following assumption {\bf A3} which is a corollary of assumptions {\bf A1} and {\bf A2}. 

{\bf A3.} We say that $f=\sum_{i\in T}x_ig_i$ has $\ell_1$ incoherence property with parameters $D$, $V$, and $r$ if for any $A\subset T$ and any $\Lambda$ such that $A\cap \Lambda =\emptyset$, $|A|+|\Lambda| \le D$ we have for any $\{c_i\}$
\begin{equation}\label{C3}
\sum_{i\in A}|x_i| \le V|A|^r\|f_A-\sum_{i\in\Lambda}c_ig_i\|.
\end{equation}
A dictionary $\D$ has $\ell_1$ incoherence property with parameters $K$, $D$, $V$, and $r$ if for any $A\subset B$, $|A|\le K$, $|B|\le D$ we have for any $\{c_i\}_{i\in B}$
$$
\sum_{i\in A} |c_i| \le V|A|^r\|\sum_{i\in B} c_ig_i\|.
$$

It is clear that {\bf A1} and {\bf A2} imply {\bf A3} with $V=C_1U$. Also, {\bf A3} implies {\bf A1} with $C_1=V$ and {\bf A2} with $U=VK^r$. Obviously, we can restrict ourselves to $r\le 1$. 

We now proceed to main results of this paper on the WCGA with respect to redundant dictionaries. The following Theorem \ref{T2.1} in the case $q=2$ was proved in \cite{LivTem}. 

 \begin{Theorem}\label{T2.1} Let $X$ be a Banach space with $\rho(u)\le \gamma u^q$, $1<q\le 2$. Suppose $K$-sparse $f^\e$ satisfies {\bf A1}, {\bf A2} and $\|f_0-f^\e\|\le \e$. Then the WCGA with weakness parameter $t$ applied to $f_0$ provides
$$
\|f_{C(t,\gamma,C_1)U^{q'}\ln (U+1) K^{rq'}}\| \le C\e\quad\text{for}\quad K+C(t,\gamma,C_1)U^{q'}\ln (U+1) K^{rq'}\le D
$$
with an absolute constant $C$.
\end{Theorem}

 It was pointed out in \cite{LivTem} that Theorem \ref{T2.1} provides a corollary for Hilbert spaces that gives sufficient conditions somewhat weaker than the known RIP conditions on $\D$ for the Lebesgue-type inequality to hold. We formulate the corresponding definitions and results. 
  Let $\D$ be the Riesz dictionary with depth $D$ and parameter $\delta\in (0,1)$. This class of dictionaries is a generalization of the class of classical Riesz bases. We give a definition in a general Hilbert space (see \cite{Tbook}, p. 306).
\begin{Definition}\label{D3.1} A dictionary $\D$ is called the Riesz dictionary with depth $D$ and parameter $\delta \in (0,1)$ if, for any $D$ distinct elements $e_1,\dots,e_D$ of the dictionary and any coefficients $a=(a_1,\dots,a_D)$, we have
\begin{equation}\label{3.3}
(1-\de)\|a\|_2^2 \le \|\sum_{i=1}^D a_ie_i\|^2\le(1+\de)\|a\|_2^2.
\end{equation}
We denote the class of Riesz dictionaries with depth $D$ and parameter $\delta \in (0,1)$ by $R(D,\de)$.
\end{Definition}
The term Riesz dictionary with depth $D$ and parameter $\delta \in (0,1)$ is another name for a dictionary satisfying the Restricted Isometry Property (RIP) with parameters $D$ and $\de$. The following simple lemma holds.
\begin{Lemma}\label{L3.1} Let $\D\in R(D,\de)$ and let $e_j\in \D$, $j=1,\dots, s$. For $f=\sum_{i=1}^s a_ie_i$ and $A \subset \{1,\dots,s\}$ denote
$$
S_A(f) := \sum_{i\in A} a_ie_i.
$$
If $s\le D$ then
$$
\|S_A(f)\|^2 \le (1+\de)(1-\de)^{-1} \|f\|^2.
$$
\end{Lemma}

Lemma \ref{L3.1} implies that if $\D\in R(D,\de)$ then it is $(D,D)$-unconditional with a constant $U=(1+\de)^{1/2}(1-\de)^{-1/2}$.

 \begin{Theorem}\label{T2.2} Let $X$ be a Hilbert space. Suppose $K$-sparse $f^\e$ satisfies  {\bf A2} and $\|f_0-f^\e\|\le \e$. Then the WOMP with weakness parameter $t$ applied to $f_0$ provides
$$
\|f_{C(t,U) K}\| \le C\e\quad\text{for}\quad K+C(t,U) K\le D
$$
with an absolute constant $C$.
\end{Theorem}
Theorem \ref{T2.2} implies the following corollaries.
 \begin{Corollary}\label{C2.1} Let $X$ be a Hilbert space. Suppose any $K$-sparse $f$ satisfies   {\bf A2}.   Then the WOMP with weakness parameter $t$ applied to $f_0$ provides
$$
\|f_{C(t,U) K}\| \le C\sigma_K(f_0,\D)\quad\text{for}\quad K+C(t,U) K\le D
$$
with an absolute constant $C$.
\end{Corollary}

 \begin{Corollary}\label{C2.2} Let $X$ be a Hilbert space. Suppose $\D\in R(D,\de)$.     Then the WOMP with weakness parameter $t$ applied to $f_0$ provides
$$
\|f_{C(t,\de) K}\| \le C\sigma_K(f_0,\D)\quad\text{for}\quad K+C(t,\de) K\le D
$$
with an absolute constant $C$.
\end{Corollary}

  We   emphasized in \cite{LivTem} that in Theorem \ref{T2.1} we impose our conditions on an individual function $f^\e$. It may happen that the dictionary does not have the Nikol'skii  $\ell_1X$ property and $(K,D)$-unconditionality but the given $f_0$ can be approximated by $f^\e$ which does satisfy assumptions {\bf A1} and {\bf A2}. Even in the case of a Hilbert space the above results from \cite{LivTem} add something new to the study based on the RIP property of a dictionary. First of all, Theorem \ref{T2.2} shows that it is sufficient to impose assumption {\bf A2} on $f^\e$ in order to obtain exact recovery and the Lebesgue-type inequality results. Second, Corollary \ref{C2.1} shows that the condition {\bf A2}, which is weaker than the RIP condition, is sufficient for exact recovery and the Lebesgue-type inequality results. Third, Corollary \ref{C2.2} shows that even if we impose our assumptions in terms of RIP we do not need to assume that $\de < \de_0$. In fact, the result works for all $\de<1$ with parameters depending on $\de$.
  
Theorem \ref{T2.1} follows from the combination of Theorems \ref{T2.3} and \ref{T2.4}.
In case $q=2$ these theorems were proved in \cite{LivTem}. 

\begin{Theorem}\label{T2.3} Let $X$ be a Banach space with $\rho(u)\le \gamma u^q$, $1<q\le 2$. Suppose for a given $f_0$ we have $\|f_0-f^\e\|\le \e$ with $K$-sparse $f:=f^\e$ satisfying {\bf A3}. Then for any $k\ge 0$ we have for $K+m \le D$
$$
\|f_m\| \le \|f_k\|\exp\left(-\frac{c_1(m-k)}{K^{rq'}}\right) +2\e, \quad q':=\frac{q}{q-1},
$$
where $c_1:= \frac{t^{q'}}{2(16\gamma)^{\frac{1}{q-1}} V^{q'}}$.
\end{Theorem}
In all theorems that follow we assume $rq'\ge 1$.
\begin{Theorem}\label{T2.4} Let $X$ be a Banach space with $\rho(u)\le \gamma u^q$, $1<q\le 2$. Suppose $K$-sparse $f^\e$ satisfies {\bf A1}, {\bf A2} and $\|f_0-f^\e\|\le \e$. Then the WCGA with weakness parameter $t$ applied to $f_0$ provides
$$
\|f_{C'U^{q'}\ln (U+1) K^{rq'}}\| \le CU\e\quad\text{for}\quad K+C'U^{q'}\ln (U+1) K^{rq'}\le D
$$
with an absolute constant $C$ and $C' = C_2(q)\gamma^{\frac{1}{q-1}} C_1^{q'} t^{-q'}$.
\end{Theorem}
We formulate an immediate corollary of Theorem \ref{T2.4} with $\e=0$.
\begin{Corollary}\label{C2.3} Let $X$ be a Banach space with $\rho(u)\le \gamma u^q$. Suppose $K$-sparse $f$ satisfies {\bf A1}, {\bf A2}. Then the WCGA with weakness parameter $t$ applied to $f$ recovers it exactly after $C'U^{q'}\ln (U+1) K^{rq'}$ iterations under condition $K+C'U^{q'}\ln (U+1) K^{rq'}\le D$.
\end{Corollary}

We formulate versions of Theorem \ref{T2.4} with assumptions {\bf A1}, {\bf A2} replaced by a single assumption {\bf A3} and replaced by two assumptions {\bf A2} and {\bf A3}. The corresponding modifications in the proofs go as in the proof of Theorem \ref{T2.3}. 

\begin{Theorem}\label{T2.5} Let $X$ be a Banach space with $\rho(u)\le \gamma u^q$, $1<q\le 2$. Suppose $K$-sparse $f^\e$ satisfies {\bf A3}  and $\|f_0-f^\e\|\le \e$. Then the WCGA with weakness parameter $t$ applied to $f_0$ provides
$$
\|f_{C(t,\gamma,q)V^{q'}\ln (VK) K^{rq'}}\| \le CVK^r\e\quad\text{for}\quad K+C(t,\gamma,q)V^{q'}\ln (VK) K^{rq'}\le D
$$
with an absolute constant $C$ and $C(t,\gamma,q) = C_2(q)\gamma^{\frac{1}{q-1}}  t^{-q'}$.
\end{Theorem}

\begin{Theorem}\label{T2.6} Let $X$ be a Banach space with $\rho(u)\le \gamma u^q$, $1<q\le 2$. Suppose $K$-sparse $f^\e$ satisfies {\bf A2}, {\bf A3}  and $\|f_0-f^\e\|\le \e$. Then the WCGA with weakness parameter $t$ applied to $f_0$ provides
$$
\|f_{C(t,\gamma,q)V^{q'}\ln (U+1) K^{rq'}}\| \le CU\e\quad\text{for}\quad K+C(t,\gamma,q)V^{q'}\ln (U+1) K^{rq'}\le D
$$
with an absolute constant $C$ and $C(t,\gamma,q) = C_2(q)\gamma^{\frac{1}{q-1}}  t^{-q'}$.
\end{Theorem}

Theorems \ref{T2.5} and \ref{T2.3} imply the following analog of Theorem \ref{T2.1}.
  
  \begin{Theorem}\label{T2.7} Let $X$ be a Banach space with $\rho(u)\le \gamma u^q$, $1<q\le 2$. Suppose $K$-sparse $f^\e$ satisfies {\bf A3}  and $\|f_0-f^\e\|\le \e$. Then the WCGA with weakness parameter $t$ applied to $f_0$ provides
$$
\|f_{C(t,\gamma,q)V^{q'}\ln (VK) K^{rq'}}\| \le C\e\quad\text{for}\quad K+C(t,\gamma,q)V^{q'}\ln (VK) K^{rq'}\le D
$$
with an absolute constant $C$ and $C(t,\gamma,q) = C_2(q)\gamma^{\frac{1}{q-1}}  t^{-q'}$.
\end{Theorem}

The following edition of Theorems \ref{T2.1} and \ref{T2.7} is also useful in applications. It follows from Theorems \ref{T2.6} and \ref{T2.3}.

 \begin{Theorem}\label{T2.8} Let $X$ be a Banach space with $\rho(u)\le \gamma u^q$, $1<q\le 2$. Suppose $K$-sparse $f^\e$ satisfies {\bf A2}, {\bf A3}  and $\|f_0-f^\e\|\le \e$. Then the WCGA with weakness parameter $t$ applied to $f_0$ provides
$$
\|f_{C(t,\gamma,q)V^{q'}\ln (U+1) K^{rq'}}\| \le C\e\quad\text{for}\quad K+C(t,\gamma,q)V^{q'}\ln (U+1) K^{rq'}\le D
$$
with an absolute constant $C$ and $C(t,\gamma,q) = C_2(q)\gamma^{\frac{1}{q-1}}  t^{-q'}$.
\end{Theorem}
 
 \section{Proofs}

 We begin with a proof of Theorem \ref{T2.3}.
 \begin{proof}  Let
$$
 f:=f^\e=\sum_{i\in T}x_ig_i,\quad |T|=K,\quad g_i\in \D.
 $$
 Denote by $T^m$ the set of indices of $g_j\in D$ picked by the WCGA after $m$ iterations, $\Gamma^m := T\setminus T^m$.
 Denote by $A_1(\D)$ the closure in $X$ of the convex hull of the symmetrized dictionary $\D^\pm:=\{\pm g,g\in D\}$.   We will bound  $\|f_m\|$ from above.  Assume $\|f_{m-1}\|\ge \e$. Let $m>k$. We bound from below
 $$
 S_m:=\sup_{\phi\in A_1(\D)} |F_{f_{m-1}}(\phi)|.
 $$
 Denote $A_m:=\Gamma^{m-1}$. Then
 $$
 S_m \ge F_{f_{m-1}}(f_{A_m}/\|f_{A_m}\|_1),
 $$
 where $\|f_A\|_1 := \sum_{i\in A} |x_i|$. Next, by Lemma 6.9, p. 342, from \cite{Tbook} we obtain
 $$
 F_{f_{m-1}}(f_{A_m}) = F_{f_{m-1}}(f^\e) \ge \|f_{m-1}\|-\e.
 $$
 Thus
 \begin{equation}\label{3.4}
 S_m\ge \|f_{A_m}\|^{-1}_1(\|f_{m-1}\|-\e).
 \end{equation}
 From the definition of the modulus of smoothness we have for any $\la$
\begin{equation}\label{3.4'}
\|f_{m-1}-\la \varphi_m \| +\|f_{m-1}+\la \varphi_m \| \le 2\|f_{m-1}\|\left(1+\rho\left(\frac{\la}{\|f_{m-1}\|}\right)\right)
\end{equation}
and by (1) from the definition of the WCGA and Lemma 6.10 from \cite{Tbook}, p. 343, we get
$$
|F_{f_{m-1}}(\varphi_m)| \ge t \sup_{g\in \D} |F_{f_{m-1}}(g)| =
$$
$$
t\sup_{\phi\in A_1(\D)} |F_{f_{m-1}}(\phi)|=tS_m.
$$
Then either $F_{f_{m-1}}(\varphi_m)\ge tS_m$ or $F_{f_{m-1}}(-\varphi_m)\ge tS_m$. Both cases are treated in the same way. We demonstrate the case $F_{f_{m-1}}(\varphi_m)\ge tS_m$. We have for $\la \ge 0$
$$
\|f_{m-1}+\la \varphi_m\| \ge F_{f_{m-1}}(f_{m-1}+\la\varphi_m)\ge  \|f_{m-1}\| + \la tS_m.
$$
From here and from (\ref{3.4'}) we obtain
 $$
 \|f_m\| \le \|f_{m-1}-\la \varphi_m\| \le \|f_{m-1}\| +\inf_{\la\ge 0} (-\la t S_m + 2\|f_{m-1}\|\rho(\la/\|f_{m-1}\|)).
 $$ We discuss here the case $\rho(u) \le \gamma u^q$. Using (\ref{3.4}) we get
 $$
 \|f_m\| \le \|f_{m-1}\|\left(1-\frac{\la t}{\|f_{A_m}\|_1} +2\gamma \frac{\la^q}{\|f_{m-1}\|^q}\right) + \frac{\e \la t}{\|f_{A_m}\|_1}.
 $$
 Let $\la_1$ be a solution of
 $$
 \frac{\la t}{2\|f_{A_m}\|_1} = 2\gamma \frac{\la^q}{\|f_{m-1}\|^q},\quad \la_1 = \left(\frac{t\|f_{m-1}\|^q}{4\gamma  \|f_{A_m}\|_1}\right)^{\frac{1}{q-1}}.
 $$
 Our assumption (\ref{C3}) gives
 \begin{eqnarray*}
 \|f_{A_m}\|_1&=& \|(f^\e-G_{m-1})_{A_m}\|_1 \le VK^r\|f^\e-G_{m-1}\|\\
 &\le&VK^r(\|f_0-G_{m-1}\|+\|f_0-f^\e\|) \le VK^r(\|f_{m-1}\|+\e).
 \end{eqnarray*}
 Specify
 $$
 \la = \left(\frac{t \|f_{A_m}\|_1^{q-1}}{16\gamma (VK^{r})^q}\right)^{\frac{1}{q-1}}.
 $$
 Then, using $\|f_{m-1}\| \ge \e$ we get
 $$
 \left(\frac{\la}{\la_1}\right)^{q-1} = \frac{\|f_{A_m}\|^q_1}{4\|f_{m-1}\|^q (VK^{r})^q} \le 1
 $$
  and  obtain
 $$
 \|f_m\| \le \|f_{m-1}\| \left( 1- \frac{t^{q'}}{2(16\gamma)^{\frac{1}{q-1}} (VK^r)^{q'}}\right) + \frac{\e t^{q'}}{(16\gamma)^{\frac{1}{q-1}} (VK^r)^{q'}}.
 $$
 Denote $c_1:= \frac{t^{q'}}{2(16\gamma)^{\frac{1}{q-1}} V^{q'}}$. Then
 $$
 \|f_m\| \le \|f_{k}\|\exp\left(-\frac{c_1(m-k)}{K^{rq'}}\right) + 2\e.
 $$
\end{proof}
We proceed to a proof of Theorem \ref{T2.4}. Modifications of this proof which are in a style of the above proof of Theorem \ref{T2.3} give Theorems \ref{T2.5} and \ref{T2.6}.
  \begin{proof} We use the above notations
   $T^m $ and  $\Gamma^m := T\setminus T^m$. Let $k\ge 0$ be fixed. Suppose
 $$
 2^{n-1} <|\Gamma^k| \le 2^n.
 $$
 For $j=1,2,\dots,n,n+1$ consider the following pairs of sets $A_j,B_j$: $A_{n+1}=\Gamma^k$, $B_{n+1}=\emptyset$; for $j\le n$, $A_j:=\Gamma^k\setminus B_j$ with $B_j\subset \Gamma^k$ is such that $|B_j|\ge |\Gamma^k|-2^{j-1}$ and for any set $J\subset \Gamma^k$ with $|J|\ge |\Gamma^k|-2^{j-1}$ we have
 $$
 \|f_{B_j}\| \le \|f_J\|.
 $$
 We note that this implies that if for some $Q\subset \Gamma^k$ we have
 \begin{equation}\label{3.5}
 \|f_Q\| <\|f_{B_j}\|\quad \text{then}\quad |Q|< |\Gamma^k|-2^{j-1}.
 \end{equation}

 For a given $b> 1$, to be specified later, denote by $L$ the index such that $(B_0:=\Gamma^k)$
 $$
 \|f_{B_0}\| < b\|f_{B_1}\|,
 $$
 $$
 \|f_{B_1}\| < b\|f_{B_2}\|,
 $$
 $$
 \dots
 $$
 $$
 \|f_{B_{L-2}}\| < b\|f_{B_{L-1}}\|,
 $$
 $$
 \|f_{B_{L-1}}\| \ge b\|f_{B_{L}}\|.
 $$
 Then
 \begin{equation}\label{3.6}
 \|f_{B_j}\| \le b^{L-1-j}\|f_{B_{L-1}}\|, \quad j=1,2,\dots,L.
 \end{equation}

  We now proceed to a general step. Let $m>k$ and let $A,B \subset \Gamma ^k$ be such that
 $ A=\Gamma^k\setminus B$. As above we bound $S_m$ from below. It is clear that $S_m\ge 0$.
 Denote $A_m := A\cap \Gamma^{m-1}$. Then
 $$
 S_m \ge F_{f_{m-1}}(f_{A_m}/\|f_{A_m}\|_1).
 $$
   Next,
 $$
 F_{f_{m-1}}(f_{A_m}) = F_{f_{m-1}}(f_{A_m}+f_B-f_B).
 $$
 Then $f_{A_m}+f_B=f^\e - f_\Lambda$ with $F_{f_{m-1}}(f_\Lambda)=0$. Moreover, it is easy to see that $F_{f_{m-1}}(f^\e)\ge\|f_{m-1}\|-\e$. Therefore,
 $$
 F_{f_{m-1}}(f_{A_m}+f_B-f_B)\ge \|f_{m-1}\|-\e - \|f_B\|.
 $$
 Thus
 $$
 S_m\ge \|f_{A_m}\|^{-1}_1\max(0,\|f_{m-1}\|-\e-\|f_B\|).
 $$
 By (\ref{C1}) we get
 $$
 \|f_{A_m}\|_1 \le C_1 |A_m|^{r}\|f_{A_m}\| \le C_1 |A|^{r}\|f_{A_m}\| .
 $$
 Then
 \begin{equation}\label{3.7}
 S_m \ge \frac{\|f_{m-1}\|-\|f_B\|-\e}{C_1 |A|^{r} \|f_{A_m}\|}.
 \end{equation}
 From the definition of the modulus of smoothness we have for any $\la$
$$
\|f_{m-1}-\la \varphi_m \| +\|f_{m-1}+\la \varphi_m \| \le 2\|f_{m-1}\|(1+\rho(\frac{\la}{\|f_{m-1}\|}))
$$
and by (1) from the definition of the WCGA and Lemma 6.10 from \cite{Tbook}, p. 343, we get
$$
|F_{f_{m-1}}(\varphi_m)| \ge t \sup_{g\in \D} |F_{f_{m-1}}(g)| =
$$
$$
t\sup_{\phi\in A_1(\D)} |F_{f_{m-1}}(\phi)|.
$$
 From here  we obtain
 $$
 \|f_m\| \le \|f_{m-1}\| +\inf_{\la\ge 0} (-\la t S_m + 2\|f_{m-1}\|\rho(\la/\|f_{m-1}\|)).
 $$
 We discuss here the case $\rho(u) \le \gamma u^q$. Using (\ref{3.7}) we get
 $$
 \|f_m\| \le \|f_{m-1}\|\left(1-\frac{\la t}{C_1|A|^{r}\|f_{A_m}\|} +2\gamma \frac{\la^q}{\|f_{m-1}\|^q}\right) + \frac{\la t(\|f_B\|+\e)}{C_1 |A|^{r} \|f_{A_m}\|}.
 $$
 Let $\la_1$ be a solution of
 $$
 \frac{\la t}{2C_1|A|^{r}\|f_{A_m}\|} = 2\gamma \frac{\la^q}{\|f_{m-1}\|^q},\quad \la_1 = \left(\frac{t\|f_{m-1}\|^q}{4\gamma C_1 |A|^{r}\|f_{A_m}\|}\right)^{\frac{1}{q-1}}.
 $$
 Our assumption (\ref{C2}) gives
 $$
 \|f_{A_m}\| \le U(\|f_{m-1}\|+\e).
 $$
 Specify
 $$
 \la = \left(\frac{t \|f_{A_m}\|^{q-1}}{16\gamma C_1 |A|^{r} U^q}\right)^{\frac{1}{q-1}}.
 $$
 Then $\la \le \la_1$ and we obtain
 \begin{equation}\label{3.8}
 \|f_m\| \le \|f_{m-1}\| \left( 1- \frac{t^{q'}}{2(16\gamma)^{\frac{1}{q-1}} (C_1 U |A|^{r})^{q'}}\right)+ \frac{ t^{q'} (\|f_B\|+\e)}{(16\gamma)^{\frac{1}{q-1}} (C_1 |A|^{r} U)^{q'}}.
 \end{equation}
 Denote $c_1:= \frac{t^{q'}}{2(16\gamma)^{\frac{1}{q-1}} (C_1 U)^{q'}}$.
 This implies for $m_2>m_1\ge k$
 \begin{equation}\label{3.9}
 \|f_{m_2}\| \le \|f_{m_1}\| (1-c_1/|A|^{rq'})^{m_2-m_1} + \frac{2c_1(m_2-m_1)}{|A|^{2r}}(\|f_B\|+\e).
 \end{equation}
 Define $m_0:= k$ and, inductively,
 $$
 m_j=m_{j-1} + \beta |A_j|^{rq'},\quad j=1,\dots,n.
 $$
 At iterations from $m_{j-1}+1$ to $m_j$ we use $A=A_j$ and obtain from (\ref{3.8})
 $$
 \|f_{m_j}\| \le \|f_{m_{j-1}}\|e^{-c_1\beta}+ 2 (\|f_{B_j}\|+\e).
 $$
 We continue it up to $j=L$. Denote $\eta:=e^{-c_1\beta}$. Then
 $$
 \|f_{m_L}\| \le \|f_k\|\eta^L +2 \sum_{j=1}^L (\|f_{B_j}\|+\e)\eta^{L-j}.
 $$
 We bound the $\|f_k\|$. It follows from the definition of $f_k$ that $\|f_k\|$ is the error of best approximation of $f_0$ by the subspace $\Phi_k$. Representing $f_0=f+f_0-f$ we see that $\|f_k\|$ is not greater than  the error of best approximation of $f$ by the subspace $\Phi_k$ plus $\|f_0-f\|$. This implies $\|f_k\|\le \|f_{B_0}\| +\e$. Therefore we continue
 $$
 \le (\|f_{B_0}\|+\e)\eta^L +2\sum_{j=1}^L (\|f_{B_{L-1}}\|(\eta b)^{L-j} b^{-1} + \e\eta^{L-j})
 $$
 $$
 \le b^{-1}\|f_{B_{L-1}}\|\left( (\eta b)^L + 2 \sum_{j=1}^L (\eta b)^{L-j}\right) + \frac{2\e}{1-\eta}.
 $$
We will specify $\beta$ later. However, we note that it will be chosen in such a way that guarantees $\eta< 1/2$. Choose $b=\frac{1}{2\eta}$. Then
 \begin{equation}\label{3.10}
 \|f_{m_L}\| \le \|f_{B_{L-1}}\|8  e^{-c_1\beta}+4\e.
 \end{equation}

 By (\ref{C2}) we get
 $$
 \|f_{\Gamma^{m_L}}\| \le U(\|f_{m_L}\|+\e) \le U(\|f_{B_{L-1}}\|8 e^{-c_1\beta} + 5\e).
 $$
 We note that in the proof of Theorem \ref{T2.5} we use the above inequality with $U=VK^r \le VK$.
 If $\|f_{B_{L-1}}\|\le 10U \e$ then by (\ref{3.10})
 $$
 \|f_{m_L}\| \le CU\e.
 $$
 If $\|f_{B_{L-1}}\|\ge 10U\e$ then
 making $\beta$ sufficiently large to satisfy $16U e^{-c_1\beta }<1$ so that $\beta = \frac{C_3\ln (U+1)}{c_1}$,  we get
 $$
 U(\|f_{B_{L-1}}\|8e^{-c_1\beta} +5\e)< \|f_{B_{L-1}}\|
 $$
 and therefore
 $$
 \|f_{\Gamma^{m_L}}\|<\|f_{B_{L-1}}\|.
 $$
 This implies
 $$
 |\Gamma^{m_L}| < |\Gamma^k| - 2^{L-2}.
 $$
   We begin with $f_0$ and apply the above argument (with $k=0$). As a result we either get the required inequality or we reduce the cardinality of support of $f$ from $|T|=K$ to $|\Gamma^{m_{L_1}}|<|T|-2^{L_1-2}$, $m_{L_1}\le \beta 2^{aL_1}$, $a:=rq'$. We continue the process and build a sequence $m_{L_j}$ such that $m_{L_j}\le \beta 2^{aL_j}$ and after $m_{L_j}$ iterations we reduce the support by at least $2^{L_j-2}$. We also note that $m_{L_j}\le \beta 2^{a} K^{a}$. We continue this process till the following inequality is satisfied for the first time
 \begin{equation}\label{3.11}
  m_{L_1}+\dots+m_{L_n}  \ge 2^{2a}\beta K^{a}.
 \end{equation}
 Then, clearly,
 $$
  m_{L_1}+\dots+m_{L_n} \le 2^{2a+1}\beta  K^{a}.
 $$
 Using the inequality
 $$
 (a_1+\cdots +a_n)^{\theta} \le a_1^\theta+\cdots +a_n^\theta,\quad a_j\ge 0,\quad \theta\in (0,1]
 $$
 we derive from (\ref{3.11}) 
 $$
 2^{L_1-2}+\dots+2^{L_n-2} \ge \left(2^{a(L_1-2)}+\dots+2^{a(L_n-2)}\right)^{\frac{1}{a}}
 $$
 $$
 \ge
 2^{-2}\left(2^{aL_1}+\dots+2^{aL_n}\right)^{\frac{1}{a}}
 $$
 $$
 \ge 2^{-2}\left((\beta)^{-1}(m_{L_1}+\dots+m_{L_n})\right)^{\frac{1}{a}}\ge K.
 $$
 Thus, after not more than $N:=2^{2a+1}\beta  K^{a}$ iterations we recover $f$ exactly and then $\|f_N\| \le \|f_0-f\|\le \e$.

 \end{proof}

 \section{Examples}

In this section we discuss applications of Theorems from Section 2 for specific dictionaries $\D$. Mostly, $\D$ will be a basis $\Psi$ for $X$. Because of that we use $m$ instead of $K$ in the notation of sparse approximation.  In some of our examples we take $X=L_p$, $2\le p<\infty$. Then it is known that $\rho(u) \le \gamma u^2$ with $\gamma = (p-1)/2$. In some other examples we take $X=L_p$, $1<p\le 2$. 
Then it is known that $\rho(u) \le \gamma u^{p}$,  with $\gamma = 1/p$.

{\bf Example 1.} Let $\Psi$ be a uniformly bounded orthogonal system normalized in $L_p(\Omega)$, $2\le p<\infty$, $\Omega$ is a bounded domain. Then we have
$$
C_1(\Omega,p)\|\psi_j\|_2\le \|\psi_j\|_p \le C_2(\Omega,p)\|\psi_j\|_2,\quad j=1,2\dots.
$$
Next, for $f=\sum_ic_i(f)\psi_i$
$$
\sum_{i\in A}|c_i(f)| = \<f,\sum_{i\in A}(\sign c_i(f))\psi_i \|\psi_i\|_2^{-2}\> 
$$
$$
\le \|f\|_2 \|\sum_{i\in A}(\sign c_i(f))\psi_i \|\psi_i\|_2^{-2}\|_2 \le C_3(\Omega,p)|A|^{1/2}\|f\|_p.
$$
Therefore $\Psi$ satisfies {\bf A3} with $D=\infty$, $V=C_3(\Omega,p)$, $r=1/2$. Theorem \ref{T2.7} gives
 \begin{equation}\label{4.4}
\|f_{C(t,p,D)m\ln (m+1)}\|_p \le C\sigma_m(f_0,\Psi)_p .
\end{equation}

{\bf Example 1q.} Let $\Psi$ be a uniformly bounded orthogonal system normalized in $L_p(\Omega)$, $1< p\le 2$, $\Omega$ is a bounded domain. Then we have
$$
C_1(\Omega,p)\|\psi_j\|_2\le \|\psi_j\|_p \le C_2(\Omega,p)\|\psi_j\|_2,\quad j=1,2\dots.
$$
Next, for $f=\sum_ic_i(f)\psi_i$
$$
\sum_{i\in A}|c_i(f)| = \<f,\sum_{i\in A}(\sign c_i(f))\psi_i \|\psi_i\|_2^{-2}\> 
$$
$$
\le \|f\|_p \|\sum_{i\in A}(\sign c_i(f))\psi_i \|\psi_i\|_2^{-2}\|_{p'} \le C_4(\Omega,p)|A|^{1-1/p'}\|f\|_p.
$$
Therefore $\Psi$ satisfies {\bf A3} with $D=\infty$, $V=C_4(\Omega,p)$, $r=1-1/p'$. Theorem \ref{T2.7} gives
 \begin{equation}\label{4.4q}
\|f_{C(t,p,D)m^{p'-1}\ln (m+1)}\|_p \le C\sigma_m(f_0,\Psi)_p .
\end{equation}

{\bf Example 2.} Let $\Psi$ be the normalized in $L_p$, $2\le p<\infty$, real $d$-variate trigonometric
system. Then Example 1 applies and gives  
for any $f_0\in L_p$
\begin{equation}\label{4.1}
\|f_{C(t,p,d)m\ln (m+1)}\|_p \le C\sigma_m(f_0,\Psi)_p .
\end{equation}
We note that (\ref{4.1}) provides some progress in Open Problem 7.1 (p. 91) from \cite{Tsurv}. 

{\bf Example 2q.} Let $\Psi$ be the normalized in $L_p$, $1<p\le 2$, real $d$-variate trigonometric
system. Then Example 1q applies and gives  
for any $f_0\in L_p$
\begin{equation}\label{4.1q}
\|f_{C(t,p,d)m^{p'-1}\ln (m+1)}\|_p \le C\sigma_m(f_0,\Psi)_p .
\end{equation}

 We need the concept of cotype of a Banach space $X$. We say that $X$ has cotype $s$ if for any finite number of elements $u_i\in X$ we have the inequality
$$
\left(\text{Average}_{\pm}\|\sum_{i}\pm u_i \|^s\right)^{1/s} \ge C_s\left(\sum_{i}\|u_i\|^s\right)^{1/s}.
$$
It is known that the $L_p$ spaces with $2\le p<\infty$ have cotype $s=p$ and $L_p$ spaces with $1< p\le 2$ have cotype $2$.
\begin{Remark}\label{R3.1} Suppose $\D$ is $(K,K)$-unconditional with a constant $U$. Assume that $X$ is of cotype $s$ with a constant $C_s$. Then $\D$ has the Nikol'skii-type $\ell_1X$ property with parameters $K,1-1/s$ and $C_1=2UC_s^{-1}$.
\end{Remark}
\begin{proof} Our assumption about $(K,K)$-unconditionality implies: for any $A$, $|A|\le K$, we have
$$
\|\sum_{i\in A} \pm x_i g_i\| \le 2U\|\sum_{i\in A} x_i g_i\|.
$$
Therefore, by $s$-cotype assumption
$$
\|\sum_{i\in A} x_i g_i\|^s \ge (2U)^{-s}C_s^s\sum_{i\in A}|x_i|^s.
$$
This implies
$$
\sum_{i\in A} |x_i| \le |A|^{1-1/s}\left(\sum_{i\in A}|x_i|^s\right)^{1/s} \le 2UC_s^{-1}|A|^{1-1/s}\|\sum_{i\in A} x_i g_i\|.
$$
\end{proof}

{\bf Example 3.} Let $X$ be a Banach space with $\rho(u)\le \gamma u^q$, $1<q\le 2$ and with cotype $s$. 
Let $\Psi$ be a normalized in $X$ unconditional basis for $X$. Then $U\le C(X,\Psi)$ and $\Psi$ satisfies {\bf A2} with $D=\infty$ and any $K$. 

By Remark \ref{R3.1} $\Psi$ satisfies {\bf A1} with $r=1-\frac{1}{s}$. Theorem \ref{T2.4} gives
\begin{equation}\label{4.3q}
\|f_{C(t,X,\Psi)m^{(1-1/s)q'}}\| \le C\sigma_m(f_0,\Psi) .
\end{equation}

{\bf Example 4.} Let $\Psi$ be the normalized in $L_p$, $2\le p<\infty$, multivariate Haar basis 
${\mathcal H}^d_p={\mathcal H}_p\times\cdots\times {\mathcal H}_p$. 
It is an unconditional basis. Also it is known that $L_p$ space with $2\le p<\infty$ has cotype $s=1/p$. Therefore, Example 3 applies in this case. We give a direct argument here. It is an unconditional basis and therefore $U\le C(p,d)$. Next, for any $A$
$$
\|\sum_{i\in A} x_i H_{i,p}\|_p \ge C(p,d)\left(\sum_{i\in A}|x_i|^{p}\right)^{1/p} \ge C(p,d)|A|^{\frac{1}{p}-1}\sum_{i\in A} |x_i|. 
$$
Therefore, we can take $r= \frac{1}{p'}$. Theorem \ref{T2.4} gives
\begin{equation}\label{4.2}
\|f_{C(t,p,d)m^{2/p'}}\|_p \le C\sigma_m(f_0,{\mathcal H}^d_p)_p .
\end{equation}
Inequality (\ref{4.2}) provides some progress in Open Problem 7.2 (p. 91) from \cite{Tsurv} in the case $2<p<\infty$. 

 {\bf Example 4q.} Let $\Psi$ be the normalized in $L_p$, $1<p\le 2$, univariate Haar basis 
${\mathcal H}_p=\{H_{I,p}\}_I$, where $H_{I,p}$ the Haar functions indexed by dyadic intervals of support of $H_{I,p}$ (we index function $1$ by $[0,1]$ and the first Haar function by $(0,1]$). Then for any finite set $A$ of dyadic intervals we have for $f=\sum_{I}c_I(f)H_{I,p}$
$$
\sum_{I\in A}|c_I| = \<f,f^*_A\>,\quad f^*_A:= \sum_{I\in A}(\sign c_I(f)) H_{I,p}\|H_{I,p}\|_2^{-2}.
$$
Therefore,
$$
\sum_{I\in A}|c_I|\le \|f\|_p\|f^*_A\|_{p'}.
$$
It is easy to check that
$$
\|H_{I,p}\|_{p'}\|H_{I,p}\|_2^{-2} = |I|^{-1/p}|I|^{1/p'}|I|^{-(1-2/p)}=1.
$$
By Lemma 1.23, p. 28, from \cite{Tbook} we get
$$
\|f^*_A\|_{p'} \le C(p)|A|^{1/p'}.
$$
Thus
$$
\sum_{I\in A}|c_I|\le C(p)|A|^{1/p'}\|f\|_p.
$$
This means that $\H_p$ satisfies {\bf A3} with $V=C(p)$ and $r=1/p'$. 
Also it is an unconditional basis and therefore satisfies {\bf A2} with $U=C(p)$. 
It is known that $L_p$ space with $1<p\le 2$ has modulus of smoothness $\rho(u)\le \gamma u^p$. Therefore, Theorem \ref{T2.8}  applies in this case and gives
\begin{equation}\label{4.2q}
\|f_{C(t,p)m}\|_p \le C\sigma_m(f_0,{\mathcal H}_p)_p .
\end{equation}
Inequality (\ref{4.2q}) solves the Open Problem 7.2 (p. 91) from \cite{Tsurv} in the case $1<p\le 2$.
 
{\bf Example 5.} Let $X$ be a Banach space with $\rho(u)\le \gamma u^2$. Assume that $\Psi$ is a normalized Schauder basis for $X$. Then for any $f=\sum_ic_i(f)\psi_i$
$$
  \sum_{i\in A}|c_i(f)| \le C(\Psi)|A|\|f\|.
$$
This implies that $\Psi$ satisfies {\bf A3} with $D=\infty$, $V=C(\Psi)$, $r=1$ and any $T$.   Theorem \ref{T2.7} gives
 \begin{equation}\label{4.5}
\|f_{C(t,X,\Psi)m^2\ln m}\| \le C\sigma_m(f_0,\Psi) .
\end{equation}
We note that the above simple argument still works if we replace the assumption that $\Psi$ is a Schauder basis by the assumption that a dictionary $\D$ is $(1,D)$-unconditional with constant $U$. Then we obtain
$$
\|f_{C(t,\gamma,U)K^2\ln K}\| \le C\sigma_K(f_0,\Psi),\quad\text{for}\quad K+C(t,\gamma,U)K^2\ln K\le D .
$$

{\bf Example 5q.} Let $X$ be a Banach space with $\rho(u)\le \gamma u^q$, $1<q\le 2$. Assume that $\Psi$ is a normalized Schauder basis for $X$. Then for any $f=\sum_ic_i(f)\psi_i$
$$
  \sum_{i\in A}|c_i(f)| \le C(\Psi)|A|\|f\|.
$$
This implies that $\Psi$ satisfies {\bf A3} with $D=\infty$, $V=C(\Psi)$, $r=1$ and any $T$.   Theorem \ref{T2.7} gives
 \begin{equation}\label{4.5q}
\|f_{C(t,X,\Psi)m^{q'}\ln m}\| \le C\sigma_m(f_0,\Psi) .
\end{equation}
We note that the above simple argument still works if we replace the assumption that $\Psi$ is a Schauder basis by the assumption that a dictionary $\D$ is $(1,D)$-unconditional with constant $U$. Then we obtain
$$
\|f_{C(t,\gamma,q,U)K^{q'}\ln K}\| \le C\sigma_K(f_0,\D),\quad\text{for}\quad K+C(t,\gamma,q,U)K^{q'}\ln K\le D .
$$

We now discuss application of Theorem \ref{T2.1} to quasi-greedy bases. We begin with a brief introduction to the theory of quasi-greedy bases. Let $X$ be an
infinite-dimensional separable Banach space with a norm
$\|\cdot\|:=\|\cdot\|_X$ and let $\Psi:=\{\psi_m
\}_{m=1}^{\infty}$ be a normalized basis for $X$.      The concept of quasi-greedy basis was introduced in \cite{KonT}.

\begin{Definition}\label{D4.1}
The basis $\Psi$ is called quasi-greedy if there exists some
constant $C$ such that
$$\sup_m \|G_m(f,\Psi)\| \leq C\|f\|.$$
\end{Definition}

Subsequently, Wojtaszczyk \cite{W1} proved that these are
precisely the bases for which the TGA merely converges, i.e.,
$$\lim_{n\rightarrow \infty}G_n(f)=f.$$

 The following lemma is from \cite{DKK} (see also \cite{DS-BT}). 

\begin{Lemma}\label{L4.1} Let $\Psi$ be a quasi-greedy basis of $X$. Then for any finite set of indices $\Lambda$ we have for all $f\in X$
$$
\|S_\Lambda(f,\Psi)\| \le C \ln(|\Lambda|+1)\|f\|.  
$$
\end{Lemma}

We now formulate a result about quasi-greedy bases in $L_p$ spaces. The following theorem is from \cite{TYY1}. We note that in the case $p=2$
 Theorem \ref{T4.1} was proved in \cite{W1}. Some notations first. For a given element $f\in X$ we consider the expansion
$$
f=\sum_{k=1}^{\infty}c_k(f)\psi_k
$$
and the decreasing rearrangement of its coefficients
$$
|c_{k_1}(f)|\ge |c_{k_2}(f)|\ge... \,\,.
$$
Denote
$$a_n(f):=|c_{k_n}(f)|.$$

\begin{Theorem}\label{T4.1} Let $\Psi=\{\psi_m\}_{m=1}^\infty$ be a quasi-greedy basis of the $L_p$ space, $1<p<\infty$. Then for each $f\in X$ we have
$$
C_1(p)\sup_n n^{1/p}a_n(f)\le  \|f\|_p\le C_2(p) \sum_{n=1}^\infty n^{-1/2}a_n(f),\quad 2\le p <\infty;
$$
$$
C_3(p)\sup_n n^{1/2}a_n(f)\le  \|f\|_p\le C_4(p) \sum_{n=1}^\infty n^{1/p-1}a_n(f),\quad 1< p \le 2.
$$
\end{Theorem}

{\bf Example 6.}  
Let $\Psi$ be a normalized   quasi-greedy basis for $L_p$, $2\le p<\infty$.   Theorem \ref{T4.1} implies for any $f=\sum_ic_i(f)\psi_i$
$$
\sum_{i\in A}|c_i(f)|\le \sum_{n=1}^{|A|} a_n(f) \le C_1(p)^{-1}\sum_{n=1}^{|A|} n^{-1/p}\|f\|_p \le C(p)|A|^{1-1/p}\|f\|_p.
$$
This means that $\Psi$ satisfies {\bf A3} with $D=\infty$, $V=C(p)$, $r=1-\frac{1}{p}$.    Theorem \ref{T2.7} gives
\begin{equation}\label{4.6}
\|f_{C(t,p)m^{2(1-1/p)} \ln (m+1)}\| \le C\sigma_m(f_0,\Psi) .
\end{equation}

{\bf Example 6q.}  
Let $\Psi$ be a normalized   quasi-greedy basis for $L_p$, $1<p\le 2$.   Theorem \ref{T4.1} implies for any $f=\sum_ic_i(f)\psi_i$
$$
\sum_{i\in A}|c_i(f)|\le \sum_{n=1}^{|A|} a_n(f) \le C_3(p)^{-1}\sum_{n=1}^{|A|} n^{-1/2}\|f\|_p \le C(p)|A|^{1/2}\|f\|_p.
$$
This means that $\Psi$ satisfies {\bf A3} with $D=\infty$, $V=C(p)$, $r=1/2$.    Theorem \ref{T2.7} gives
\begin{equation}\label{4.6q}
\|f_{C(t,p)m^{p'/2} \ln (m+1)}\| \le C\sigma_m(f_0,\Psi) .
\end{equation}

{\bf Example 7.} Let $\Psi$ be a normalized uniformly bounded orthogonal quasi-greedy basis for $L_p$, $2\le p<\infty$. For existence of such bases see \cite{N}. Then orthogonality implies that we can take $r=1/2$.  We obtain from Lemma \ref{L4.1} that $\Psi$ is $(K,\infty)$ unconditional with $U\le C\ln (K+1)$. Theorem \ref{T2.8} gives
\begin{equation}\label{4.7}
\|f_{C(t,p,\Psi)m  \ln\ln(m+3)}\|_p \le C\sigma_m(f_0,\Psi)_p .
\end{equation}

{\bf Example 7q.} Let $\Psi$ be a normalized uniformly bounded orthogonal quasi-greedy basis for $L_p$, $1<p\le 2$. For existence of such bases see \cite{N}. Then orthogonality implies that we can tame $r=1/2$.  We obtain from Lemma \ref{L4.1} that $\Psi$ is $(K,\infty)$ unconditional with  $U\le C\ln (K+1)$. Theorem \ref{T2.8} gives
\begin{equation}\label{4.7q}
\|f_{C(t,p,\Psi)m^{p'/2}  \ln\ln(m+3)}\|_p \le C\sigma_m(f_0,\Psi)_p .
\end{equation}

\section{Discussion}

We study sparse approximation. In a general setting we study an algorithm (approximation method) $\A = \{A_m(\cdot,\D)\}_{m=1}^\infty$ with respect to a given dictionary $\D$. The sequence of mappings $A_m(\cdot,\D)$ defined on $X$ satisfies the condition: for any $f\in X$, $A_m(f,\D)\in \Sigma_m(\D)$. In other words, $A_m$ provides an $m$-term approximant with respect to $\D$. It is clear that for any $f\in X$ and any $m$ we have
$$
\|f-A_m(f,\D)\| \ge \sigma_m(f,\D).
$$
We are interested in such pairs $(\D,\A)$ for which the algorithm $\A$ provides approximation close to best $m$-term approximation. We introduce the corresponding definitions.
\begin{Definition}\label{D5.1} We say that $\D$ is a greedy dictionary with respect to $\A$ if there exists a constant $C_0$ such that for any $f\in X$ we have
\begin{equation}\label{5.1}
\|f-A_m(f,\D)\| \le C_0\sigma_m(f,\D).
\end{equation}
\end{Definition}
If $\D$ is a greedy dictionary with respect to $\A$ then $\A$ provides ideal (up to a constant $C_0$) $m$-term approximations for every $f\in X$. 

\begin{Definition}\label{D5.2} We say that $\D$ is an almost greedy dictionary with respect to $\A$ if there exist two constant $C_1$ and $C_2$ such that for any $f\in X$ we have
\begin{equation}\label{5.2}
\|f-A_{C_1m}(f,\D)\| \le C_2\sigma_m(f,\D).
\end{equation}
\end{Definition}
If $\D$ is an almost greedy dictionary with respect to $\A$ then $\A$ provides almost ideal sparse approximation. It provides $C_1m$-term approximant as good (up to a constant $C_2$) as ideal $m$-term approximant for every $f\in X$. 
We also need a more general definition. Let $\phi(u)$ be a  function such that 
$\phi(u)\ge 1$. 
\begin{Definition}\label{D5.3} We say that $\D$ is a $\phi$-greedy dictionary with respect to $\A$ if there exists a constant $C_3$ such that for any $f\in X$ we have
\begin{equation}\label{5.3}
\|f-A_{\phi(m)m}(f,\D)\| \le C_3\sigma_m(f,\D).
\end{equation}
\end{Definition}

If $\D=\Psi$ is a basis then in the above definitions we replace dictionary by basis. 
In the case $\A=\{G_m(\cdot,\Psi)\}_{m=1}^\infty$ is the TGA the theory of greedy and almost greedy bases is well developed (see \cite{Tbook}). We present two results on characterization of these bases.  A basis $\Psi$ in a Banach space $X$ is called {\it
democratic} if there is a constant $C(\Psi)$  such that
\begin{equation}\label{5.4} 
\|\sum_{k\in A}\psi_k\|\le
C(\Psi)\|\sum_{k\in B}\psi_k\|
\end{equation}
 if $|A|= |B|$.
 This concept was introduced in \cite{KonT}. 
 In \cite{DKKT} we defined a democratic basis as the one satisfying (\ref{5.4}) if $|A|\le |B|$. It is known that for quasi-greedy bases the above two definitions are equivalent. 
 It was proved in \cite{KonT} (see Theorem 1.15, p. 18, \cite{Tbook}) that a basis is greedy with respect to TGA if and only if it is unconditional and democratic. It was proved in \cite{DKKT} (see Theorem 1.37, p. 38, \cite{Tbook}) that a basis is almost greedy with respect to TGA if and only if it is quasi-greedy and democratic. 
 
 Example 4q is the first result about almost greedy bases with respect to WCGA in Banach spaces.  It shows that the univariate Haar basis is an almost greedy basis with respect to the WCGA in the $L_p$ spaces for $1<p\le 2$. 
 Example 1 shows that uniformly bounded orthogonal bases are $\phi$-greedy 
 bases with respect to WCGA with $\phi(u) = C(t,p,D)\ln(u+1)$ in the $L_p$ spaces for $2\le p<\infty$. We do not know if these bases are almost greedy with respect to WCGA. They are good candidates for that. 

It is known (see \cite{Tbook}, p. 17) that the univariate Haar basis is a greedy basis with respect to TGA for all $L_p$, $1<p<\infty$. Example 4 only shows that it is a $\phi$-greedy basis with respect to WCGA with $\phi(u) = C(t,p)u^{1-2/p}$ in the $L_p$ spaces for $2\le p<\infty$. It is much weaker than the corresponding results for the $\H_p$, $1<p\le 2$, and for the trigonometric system, $2\le p<\infty$ (see Example 2). We do not know if this result on the Haar basis can be substantially improved. At the level of our today's technique we can observe that the Haar basis is ideal (greedy basis) for the TGA in $L_p$, $1<p<\infty$, almost ideal (almost greedy basis) for the WCGA in $L_p$, $1<p\le 2$, and that the trigonometric system is very good for the WCGA in $L_p$, $2\le p<\infty$. 

Example 2q  shows that our results for the trigonometric system in $L_p$, $1<p<2$, are not as strong as for $2\le p<\infty$. We do not know if it is a lack of appropriate technique or it reflects the nature of the WCGA with respect to the trigonometric system. 

We note that properties of a given basis with respect to TGA and WCGA could be very different. For instance, the class of quasi-greedy bases (with respect to TGA) is a rather narrow subset of all bases. It is close in a certain sense to the set of unconditional bases. The situation is absolutely different for the WCGA. If $X$ is uniformly smooth then WCGA converges for each $f\in X$ with respect to any dictionary in $X$. Moreover, Example 5q shows that 
if $X$ is a Banach space with $\rho(u)\le \gamma u^q$ then any basis $\Psi$ is  $\phi$-greedy with respect to WCGA with $\phi(u)=C(t,X,\Psi)u^{q'-1}\ln (u+1)$. 

It is interesting to compare Theorem \ref{T2.3} with the following known result. The following theorem provides rate of convergence (see \cite{Tbook}, p. 347). As above we denote by $A_1(\D)$ the closure in $X$ of the convex hull of the symmetrized dictionary $\D^\pm :=\{\pm g:g\in \D\}$. 
\begin{Theorem}\label{T5.1} Let $X$ be a uniformly smooth Banach space with modulus of smoothness $\rho(u)\le \gamma u^q$, $1<q\le 2$. Take a number $\e\ge 0$ and two elements $f_0$, $f^\e$ from $X$ such that
$$
\|f_0-f^\e\| \le \e,\quad
f^\e/A(\e) \in A_1(\D),
$$
with some number $A(\e)>0$.
Then, for the  WCGA    we have  
$$
\|f^{c,\tau}_m\| \le  \max\left(2\e, C(q,\gamma)(A(\e)+\e)(1+\sum_{k=1}^mt_k^{q'})^{-1/q'}\right) . 
$$
\end{Theorem}
Both   Theorem \ref{T5.1} and Theorem \ref{T2.3} provide stability of the WCGA with respect to noise. 
  In order to apply them for noisy data we interpret $f_0$ as a noisy version of a signal and $f^\e$ as a noiseless version of a signal. Then, assumption $f^\e/A(\e)\in A_1(\D)$ describes our smoothness assumption on the noiseless signal and assumption $f^\e \in \Sigma_K(\D)$ describes our structural assumption on the noiseless signal. 
  In fact, Theorem \ref{T5.1}  simultaneously takes care of two issues: noisy data and approximation in an interpolation space.
  Theorem \ref{T5.1} can be applied for approximation of $f_0$ under assumption that $f_0$ belongs  to one  
of interpolation spaces between $X$ and the space generated by the $A_1(\D)$-norm (atomic norm).   

Concluding, we briefly describe the contribution of this paper. First, we present a study of the Lebesque-type inequalities with respect to the WCGA in Banach spaces with $\rho(u) \le \gamma u^q$, $1<q\le 2$, under conditions {\bf A1} and {\bf A2}. In the case $q=2$ it has been done in \cite{LivTem}. The case $1<q<2$ uses the same ideas as in \cite{LivTem}. Second, we introduce a new condition {\bf A3} and study the WCGA with respect to dictionaries satisfying either {\bf A3} or {\bf A2} and {\bf A3}. Condition {\bf A3} and a combination of {\bf A2} and {\bf A3} turn out to be more powerful in applications than 
{\bf A1} combined with {\bf A2}. Third, we apply the general theory developed in Sections 2 and 3 for bases. Surprisingly, this technique works very well for very different bases. It provides first results on the Lebesque-type inequalities for the WCGA with respect to bases in Banach spaces. Some of these results (for the $\H_p$, $1<p\le 2$, and for the $\mathcal R \mathcal T_p$, $2\le p<\infty$) are strong. This demonstrates that the technique used is an appropriate and powerful method.

\end{document}